\documentclass[anonymous=false, %
               format=acmsmall, %
               review=false, %
               screen=true, %
               nonacm=true]{acmart}

\usepackage[ruled]{algorithm2e} 
\usepackage{url}
\usepackage{hyperref}
\usepackage{listings}
\usepackage{amsthm}
\theoremstyle{definition}
\newtheorem{definition}{Definition}
\newtheorem{example}{Example}
\usepackage{GoMono}

\newcommand{\appropto}{\mathrel{\vcenter{
  \offinterlineskip\halign{\hfil$##$\cr
    \propto\cr\noalign{\kern1pt}\sim\cr\noalign{\kern-1pt}}}}}

\lstdefinelanguage{Golang}%
  {morekeywords=[1]{package,import,func,type,struct,return,defer,panic,%
     recover,select,var,const,iota,},%
   morekeywords=[2]{string,uint,uint8,uint16,uint32,uint64,int,int8,int16,%
     int32,int64,bool,float,float32,float,complex64,complex128,byte,rune,uintptr,%
     error,interface},%
   morekeywords=[3]{map,slice,make,new,nil,len,cap,copy,close,true,false,%
     delete,append,real,imag,complex,chan,},%
   morekeywords=[4]{for,break,continue,range,go,goto,switch,case,fallthrough,if,%
     else,default,},%
   morekeywords=[5]{Println,Printf,Error,Print,},%
   sensitive=true,%
   morecomment=[l]{//},%
   morecomment=[s]{p*}{*/},%
   morestring=[b]',%
   morestring=[b]",%
   morestring=[s]{`}{`},%
}

\acmConference[]{}{}{}

\setcopyright{none}

\title{Stochastic Probabilistic Programs}
\author{David Tolpin}
\affiliation{
	\institution{Ben-Gurion University of the Negev}
	\department{Computer Science Department}
    \country{Israel}
}
\affiliation{
	\institution{PUB+}
}
\email{david.tolpin@gmail.com}

\author{Tomer Dobkin}
\affiliation{
    \institution{Ben-Gurion University of the Negev}
	\department{Computer Science Department}
    \country{Israel}
}
\email{tomer.dobkin@gmail.com}

\begin{abstract}
    We introduce the notion of a stochastic probabilistic
    program and present a reference implementation of a
    probabilistic programming facility supporting specification
    of stochastic probabilistic programs and inference in them.
    Stochastic probabilistic programs allow straightforward
    specification and efficient inference in models with
    nuisance parameters, noise, and nondeterminism. We give
    several examples of stochastic probabilistic programs, and
    compare the programs with corresponding deterministic
    probabilistic programs in terms of model specification and
    inference. We conclude with discussion of open research
    topics and related work.
\end{abstract}

\begin{document}
\maketitle

\section{Introduction}

Perhaps somewhat counterintuitively, most probabilistic programs
are deterministic. The property that distinguishes a
probabilistic program is that a probabilistic program computes
the (unnormalized) probability of its execution~\cite{WVM14}. An
execution is summarized as an instantiation of the program
\textit{trace}, and the probability is a function of the trace.
Some probabilistic programming frameworks require that the trace
shape is specified upfront~\cite{Stan17,T19}, others allow
introduction of trace components
dynamically~\cite{GMR+08,P09,GS15,GXG18,TMY+16}, in the course
of a program execution.

A probabilistic programming framework passes the probabilistic
program as an argument to an inference algorithm. The algorithm
may collect any program output for later summarization, but
otherwise ignores the output and uses only the trace probability
for inference. In some probabilistic programming languages the
locations where the probability is computed are syntactically
explicit, and inference algorithms may exploit the program
structure. For example, in Stan~\cite{Stan17} operator $\sim$ and an
increment of variable \lstinline{target} update the trace
probability; in Anglican~\cite{TMY+16}, both \lstinline{sample} and
\lstinline{observe} update the trace probability, in addition,
\lstinline{sample} adds a component to the trace; but neither
$\sim$ nor \lstinline{sample} introduce randomness in the
computation of probability.

However, stochastic computation of trace probability naturally
comes up in at least two different contexts.  In one context, nuisance
parameters must be marginalized: instead of inferring the posterior
distribution over both parameters of interest and nuisance
parameters, and marginalizing over a representation of the
posterior (for example, by summing up samples in Monte Carlo
approximations), the program can compute the execution
probability stochastically by sampling nuisance parameters, and
perform marginalization during inference. For example,
stochastic computation can be used to represent flipping a coin
with an unobserved outcome. The other context is the handling of
nondeterminism, where the posterior distribution of the trace
should be inferred under any outcome of non-deterministic
choices following a certain distribution, such as for policy
learning in stochastic domains~\cite{MPT+16}. These two contexts admit
different inference schemes, but both involve stochastic
computation of trace probability and would benefit from
consistent representation in probabilistic programs.

In this work, we formally define a stochastic probabilistic
program (Section~\ref{sec:spp}) and then introduce inference
schemes for both marginalization and nondeterminism
(Section~\ref{sec:infer}). We present a reference implementation
of probabilistic programming facility that support stochastic
probabilistic programs and show that such programs can be
supported with little extra effort compared to traditional,
deterministic probabilistic programs (Section~\ref{sec:impl}).
In case studies (Section~\ref{sec:studies}), we implement
benchmark problems from the literature as stochastic
probabilistic programs and compare both specification of the
programs and inference in them to deterministic implementations.
We also bring examples of models which are naturally represented
as stochastic probabilistic programs but are difficult to
implement without introducing stochasticity. Probabilistic
program examples in Sections~\ref{sec:spp}--\ref{sec:infer} are
in a subset of the Go programming language~\cite{Golang}, but
should be legible for readers who are not familiar with Go.

\paragraph{Contributions} This work brings the following contributions:
\begin{itemize}
	\item definition of a stochastic probabilistic program;
	\item posterior inference on stochastic probabilistic
		programs for marginalization and nondeterminism;
	\item a reference implementation of probabilistic
		programming with support for stochastic probabilistic
		programs.
\end{itemize}

\paragraph{Notation} We denote by $p(\pmb{x})$ the probability
or probability density of random variable $\pmb{x}$, and by
$p(\pmb{x}|\pmb{y})$ the conditional probability or probability
density of $\pmb{x}$ given $\pmb{y}$, depending on the domain of
$\pmb{x}$.  We write $\pmb{x} \sim p(\cdot)$ when $\pmb{x}$ is a
random variable with probability $p(\cdot)$. We denote by
$\tilde p(\cdot)$ a probability known up to a normalization
constant (unnormalized probability).

\section{A Stochastic Probabilistic Program}
\label{sec:spp}

Different definitions of (deterministic) probabilistic programs
are given in the literature and reflect different views and
accents. For the purpose of this work, let us give the following
broad definition:

\begin{definition}A deterministic probabilistic program $\Pi$
	that defines a distribution over traces $p(\pmb{x}|\pmb{y})$ is a
    computer program that accepts a \textit{trace assignment}
    $\pmb{x}$ as one of its arguments, and returns the
    unnormalized probability of $\pmb{x}$ conditioned on other
    arguments $\pmb{y}$:

    \begin{equation}
        \Pi(\pmb{x}, \pmb{y}) \Rightarrow \tilde p(\pmb{x}|\pmb{y}).
    \end{equation}
\end{definition}

The trace assignment $\pmb{x}$ may have the form of a vector, or of
a list of address-value pairs, or any other form suitable
for a particular implementation. Accordingly, let us define a
stochastic probabilistic program:

\begin{definition}A stochastic probabilistic program $\Xi$
	that defines a distribution over traces $p(\pmb{x}|\pmb{y})$ is a
    computer program that accepts a \textit{trace assignment}
    $\pmb{x}$ as one of its arguments, and returns the
	unnormalized probability of $\pmb{x}$ conditioned on other
	arguments $\pmb{y}$ and random variable $\pmb{z}$
	conditioned on $\pmb{y}$:

    \begin{align}
		\label{eqn:spp}
		\pmb{z} & \sim p(\pmb{z}|\pmb{y}) \\ \nonumber
        \Xi(\pmb{x}, \pmb{y}) & \Rightarrow \tilde p(\pmb{x}|\pmb{y},\pmb{z}).
    \end{align}
	\label{def:spp}
\end{definition}

A rationale for this definition is that $\pmb{z}$ corresponds to
nuisance parameters or nondeterministic choices inside the
program. Let us illustrate a stochastic probabilistic program on
an example: 

\begin{example}
    A survey is conducted among company's employees. The survey
    contains a single question: "Are you satisfied with your
    compensation?" To preserve employees' privacy, the employee
    flips a coin before answering the question. If the coin
    shows head, the employee answers honestly; otherwise, the
    employee flips a coin again, and answers 'yes' on head, 'no'
    on tail. Based on survey outcomes, we want to know how many
    of the employees are satisfied with their compensations.
    \label{ex:survey}
\end{example}

\begin{figure}
    \begin{lstlisting}
 1 func Survey(theta float, y []bool) float {
 2     prob := Beta(1, 1).pdf(theta)
 3     for i := 0; i != len(y); i++ {
 4         coin := Bernoulli(0.5).sample()
 5         if coin {
 6             prob *= Bernoulli(theta).cdf(y[i])
 7         } else {
 8             prob *= Bernoulli(0.5).cdf(y[i])
 9         }
10     }
11     return prob
12 }
    \end{lstlisting}
    \caption{A probabilistic program for the compensation
    survey. Nuisance variable \lstinline{coin} makes the program
    stochastic.}
    \label{fig:survey-stochastic}
\end{figure}

\paragraph{Stochastic probabilistic program} The stochastic
probabilistic program modelling the survey
(Figure~\ref{fig:survey-stochastic}) receives two parameters:
probability \lstinline{theta} that a randomly chosen employee is
satisfied with the compensation and vector \lstinline{y} of
boolean survey outcomes. The program trace consists of a single
variable \lstinline{theta}.  There are nuisance random variables
\lstinline{coin} representing point flips which are sampled
inside the program from their prior distribution, but are not
included in the trace --- this makes the probabilistic program
stochastic. The unnormalized probability of the trace is
accumulated in variable \lstinline{prob}.  First, a Beta prior
is imposed on \lstinline{theta} (line 2).  Then,
\lstinline{prob} is multiplied by the probability of each answer
given \lstinline{theta} and \lstinline{coin} (lines 5--9).

\paragraph{Manual marginalization} The program in
Figure~\ref{fig:survey-stochastic} can be rewritten as a
deterministic probabilistic program
(Figure~\ref{fig:survey-deterministic}). Instead of flipping a
coin as in line~4 of the stochastic program in
Figure~\ref{fig:survey-stochastic}, the probability of
observation \lstinline{y[i]} is computed in lines~5--6 as the
sum of probabilities given either head or tail, weighted by the
probabilities of head and tail (both are 0.5). Due to explicit
marginalization, the generative structure of the program becomes
less obvious, but one can still argue that a stochastic
probabilistic program can be straightforwardly reduced to a
deterministic one.
\begin{figure}
    \begin{lstlisting}
 1 func Survey(theta float, y []bool) float {
 2     prob := Beta(1, 1).pdf(theta)
 3     for i := 0; i != len(y); i++ {
 4         prob *= 0.5*Bernoulli(theta).cdf(y[i]) + 0.5*Bernoulli(0.5).cdf(y[i])
 5     }
 6     return prob
 7 }
    \end{lstlisting}
    \caption{A deterministic probabilistic program for the compensation
    survey. The probability is marginalized over the coin flip
    in the code of the program.}
    \label{fig:survey-deterministic}
\end{figure}
It is though easy to imagine a problem that is almost as simple
as in the example but for which manual marginalization is not
possible. Imagine that the distribution of coin flip outcomes is
not known but obtained from a black-box source. The survey still
can be conducted, as long as the source is stationary, if the
probability of each outcome is unknown and even if consequent
coin flips are not independent, by replacing line~4 in
Figure~\ref{fig:survey-stochastic} by 
\begin{lstlisting}
 4         coin := Coins.sample()
\end{lstlisting}
where \lstinline{Coins} is a black-box random source. However,
after such modification the program code cannot be marginalized,
and there is no obvious deterministic equivalent of the
stochastic probabilistic program. Worse yet, even if all
instantiations of \lstinline{coin} are included in the trace,
making the trace \lstinline{1+len(y)}-dimensional, inference
cannot be performed on the resulting deterministic probabilistic
program because the probability of a coin flip outcome with
respect to the random source \lstinline{Coins} is unknown.
Meanwhile, the stochastic probabilistic program constitutes a
valid inference model, as we show in the following sections.

\section{Inference in Stochastic Probabilistic Programs}
\label{sec:infer}

A stochastic probabilistic program of the same shape may
appear in two different contexts. In one context, like in
Example~\ref{ex:survey}, stochasticity models nuisance
parameters, such as latent noise, and the posterior distribution
of the trace is marginalized over the nuisance parameters. In
the other context, stochasticity models nondeterminism, and the
posterior distribution of the trace is inferred in expectation
over all possible outcomes of nondeterministic choices. As an
illustration of the latter context, consider the following
example:
\begin{example}
	A player throws a ball into the basket at distance $D$.
	Knowing that the initial velocity $v$ of the ball varies with a
	known distribution $D$, at which angle $\alpha$ should the
	player throw the ball to hit the basket?
    \label{ex:ball}
\end{example}

For simplicity, let's assume that the player's hands and the
basket are at the same level and that air drag can be ignored.
Then, for given $\alpha$ and $v$ the final distance $d$ of the
ball at the basket level is:
\begin{equation}
    d = v \cos\alpha \frac {2v \sin\alpha} g = \frac {v^2} g \sin 2\alpha
\end{equation}
where $g$ is the acceleration of gravity. The corresponding
stochastic probabilistic program is shown in Figure~\ref{fig:ball}.
In the program code, \lstinline{Pi} denotes constant $\pi =
3.1415926...$ and \lstinline{g} is the acceleration of gravity
$g = 9.80665$.
\begin{figure}
    \begin{lstlisting}
1 func Ball(alpha float, L float) float {
2     prob := Normal(Pi/4, Pi/8).pdf(alpha)
3     v := D.sample()
4     d := v*v/g*sin(2*alpha)
5     prob *= Normal(L, 1).pdf(d)
6     return prob
7 }
    \end{lstlisting}
    \caption{Probabilistic program for inferring the throw angle.}
    \label{fig:ball}
\end{figure}
Note that the question we want to answer with this probabilistic
program is different from just inferring the posterior
distribution of $\alpha$ conditioned on observation $d=L$, with
$v$ as a nuisance parameter (which would be the case if one
observed the ball falling at distance $L$). To model
nondeterminism, one needs to `cut' the dependency between the
distribution of $v$ and the observation, and infer $\alpha$ for
all $v$ drawn from $D$.

In the rest of the section, we describe posterior inference for
both contexts. We limit analysis to the case of $\pmb{x} \in
\mathcal{R}^n$ and $p_F(\pmb{x}|\pmb{y},\pmb{z})$ differentiable
by $\pmb{x}$. We represent the posterior inference as the
problem of drawing samples from posterior distribution
$p(\pmb{x}|\pmb{y})$ given unnormalized probability density
$\tilde p(\pmb{x}|\pmb{y},\pmb{z})$ and a random source of
$\pmb{z}$.  In either context, the inference can be based on
stochastic gradient Hamiltonian Monte Carlo
(sgHMC)~\cite{CFG14}, but differs in the way the stochastic
gradient is computed. Similarly, maximum \textit{a posteriori}
estimation can be performed using a stochastic gradient
optimization algorithm~\cite{R16}.

\subsection{Marginalization}

The unnormalized probability density $\tilde p(\pmb{x}|\pmb{y})$
is a marginalization of $\tilde p(\pmb{x}|\pmb{y},\pmb{z})$ over
$\pmb{z}$:
\begin{equation}
	\tilde p(\pmb{x}|\pmb{y}) = \int_z p(\pmb{z}|\pmb{y}) \tilde p(\pmb{x}|\pmb{y},\pmb{z})d\pmb{z}
\end{equation}
Posterior inference and maximum \textit{a posteriori} estimation
require a stochastic estimate of $\nabla_{\pmb{x}} \log \tilde p(\pmb{x}|\pmb{y})$:
\begin{align}
	\nonumber \nabla_{\pmb{x}} \log \tilde p(\pmb{x}|\pmb{y}) &= \frac {\int_z p(\pmb{z}|\pmb{y}) \tilde p(\pmb{x}|\pmb{y},\pmb{z})\nabla_x \log \tilde p(\pmb{x}|\pmb{y},\pmb{z})d\pmb{z}} {\tilde p(\pmb{x}|\pmb{y})}  \\
	\nonumber &= \int_z p(\pmb{z}|\pmb{y}) \frac {\tilde p(\pmb{x}|\pmb{y},\pmb{z})} {\tilde p(\pmb{x}|\pmb{y})} \nabla_x \log \tilde p(\pmb{x}|\pmb{y},\pmb{z})d\pmb{z} = \int_z  \frac {p(\pmb{x}|\pmb{y},\pmb{z})p(\pmb{z}|\pmb{y})} {p(\pmb{x}|\pmb{y})} \nabla_x \log \tilde p(\pmb{x}|\pmb{y},\pmb{z})d\pmb{z} \\
	&= \int_z p(\pmb{z}|\pmb{x},\pmb{y}) \nabla_x \log \tilde p(\pmb{x}|\pmb{y},\pmb{z})d\pmb{z}
\end{align}
By Monte Carlo approximation,
\begin{align}
	\label{eqn:grad-mc-marg}
	\nonumber \pmb{z}_i & \sim p(\pmb{z}|\pmb{x}, \pmb{y}) \\
	\nabla_{\pmb{x}} \log \tilde p(\pmb{x}|\pmb{y}) & \approx \frac 1 N \sum_{i=1}^N \nabla_x \log \tilde p(\pmb{x}|\pmb{y}, \pmb{z}_i)
\end{align}
Draws of $\pmb{z}_i$ are conditioned on $\pmb{x}$ and can be
approximated by a Monte Carlo method, such as Markov chain Monte
Carlo; $p(\pmb{z}|\pmb{y})$ need not be known for the
approximation. Note that there are two Monte Carlo
approximations involved: 
\begin{enumerate}
	\item gradient $\nabla_x \log \tilde
p(\pmb{x}|\pmb{y})$ is approximated by a finite sum of
gradients $\nabla_x \log \tilde p(\pmb{x}|\pmb{y}, \pmb{z_i})$;
	\item draws of $\pmb{z}_i$ from $p(\pmb{z}|\pmb{x}, \pmb{y})$ are
		approximated.
\end{enumerate}

An intuition behind estimate (\ref{eqn:grad-mc-marg}) is that
for marginalization, assignments to nuisance parameters which
make the trace assignment more likely contribute more to the
gradient estimate.

\subsection{Nondeterminism}

When a stochastic probabilistic program is used to represent
nondeterminism, the posterior inference is performed with
respect to \textit{all} possible instantiations of $\pmb{z}$ in
(\ref{eqn:spp}) (rather than \textit{any} instantiation, as in
the case of marginalization). Thus, $\tilde p(\pmb{x}|\pmb{y})$
is a geometric integral of $p(\pmb{x}|\pmb{y},\pmb{z})$ over
$\pmb{z}$:
\begin{equation}
	\tilde p(\pmb{x}|\pmb{y}) = \prod_{\pmb{z}} \tilde
	p(\pmb{x}|\pmb{y},\pmb{z})^{p(\pmb{z}|\pmb{y})}dz = \exp\left(\int_{\pmb{z}} p(\pmb{z}|\pmb{y}) \log \tilde p(\pmb{x}|\pmb{y},\pmb{z}) dz\right)
\end{equation}
Again, posterior inference and maximum \textit{a posteriori} estimation
require a stochastic estimate of $\nabla_{\pmb{x}} \log \tilde
p(\pmb{x}|\pmb{y})$:
\begin{equation}
	\nabla_x \log \tilde p(\pmb{x}|\pmb{y}) = \int_{\pmb{z}} p(\pmb{z}|\pmb{y}) \nabla_x \log \tilde p(\pmb{x}|\pmb{y},\pmb{z})d \pmb{z}
\end{equation}
By Monte Carlo approximation,
\begin{align}
	\label{eqn:grad-mc-nond}
	\nonumber \pmb{z}_i & \sim p(\pmb{z}|\pmb{y}) \\
	\nabla_{\pmb{x}}\log \tilde p(\pmb{x}|\pmb{y}) & \approx \frac 1 N \sum_{i=1}^N \nabla_{\pmb{x}} \log \tilde p(\pmb{x}|\pmb{y},\pmb{z_i})
\end{align}
Unlike in the marginalization context, $\pmb{z}_i$ are drawn
from the prior distribution of $\pmb{z}$. Here too,
$p(\pmb{z}|\pmb{y})$ need not be known. 

The case of nondeterminism bears similarity to the setting for
which sgHMC was introduced --- posterior inference on large or
streaming data, equation (5) in~\cite{CFG14}. A difference is
that in~\cite{CFG14}, the gradient scales up with the data size,
and the influence of the prior distribution of $\pmb{x}$
vanishes as the data size grows. In the nondeterminism setting,
the expectation of the gradient
estimate~(\ref{eqn:grad-mc-nond}) remains the same independently
of the number of Monte Carlo samples.

\section{Implementation}
\label{sec:impl}

We implemented inference in stochastic probabilistic programs
using Infergo~\cite{T19}. Thanks to reliance of Infergo on pure
Go code for probabilistic programs, no changes were necessary to
the way probabilistic programs are represented --- it is
sufficient just to draw samples from a random number generator
provided by one of Go libraries, for example, the standard
library or Gonum~\cite{Gonum}.

Implementations of sgHMC variants for case studies in this paper
are available at \url{http://bitbucket.org/dtolpin/spp-studies}
and will be included in a future version of Infergo.

\section{Case Studies}
\label{sec:studies}

We evaluated inference in stochastic probabilistic programs on
several models. In each evaluation, we specified two models: a
stochastic model and a corresponding deterministic model.
Inference on stochastic models was performed using either sgHMC
for nondeterminism, or MHMC, a variant of sgHMC with the gradient
computed as in (\ref{eqn:grad-mc-marg}), for marginalization;
HMC was used on deterministic models.  In each of the case
studies we checked that stochastic and deterministic models
yield consistent results. The probabilistic programs used in
case studies are available at
\url{http://bitbucket.org/dtolpin/spp-studies}. In the appendix,
we provide the source code of stochastic and deterministic
probabilistic programs for three studies: 
\begin{itemize}
	\item compensation survey (Example~\ref{ex:survey});
	\item ball throw (Example~\ref{ex:ball});
	\item Gaussian mixture model.
\end{itemize}

\section{Discussion}
\label{sec:disc}

In this paper, we introduced the notion of a stochastic
probabilistic program and inference schemes for the contexts of
marginalization and nondeterminism, in which stochastic
probabilistic programs come up. Stochastic probabilistic
programs facilitate natural specification of probabilistic
models, and support inference in the models, for which
deterministic probabilistic programs are impossible or difficult
to write. We focused our analysis on the case of differentiable
probabilistic programs with a real-valued parameter vector,
however this is not a limitation in general --- stochastic
probabilistic programs can be specified for support of any type,
albeit different inference algorithms may be required.

Our understanding of stochastic probabilistic programs, classes
of models for which they are best suited, and inference
schemes is still at an early stage and evolving. In particular,
we are working, towards future publications, on the analysis of
stochastic gradient estimate for the marginalization context.
Another research direction worth exploring is programmatic
transformation of stochastic probabilistic programs into
deterministic ones when the latter exist. Such transformation
would allow specifying models as stochastic probabilistic
programs where such representation is natural but applying
efficient and well developed inference techniques for
deterministic probabilistic programs on transformed
representations. Last but not least, we are looking into
introducing stochastic probabilistic program support into
existing probabilistic programming frameworks, through
transformation to deterministic probabilistic programs or by
extending the frameworks.

\section{Acknowledgements}

Development of Infergo is supported by PUB+, \url{http://pubplus.com/}.

\bibliography{refs}
\bibliographystyle{acm-reference-format}

\clearpage
\appendix

\section{Source code for case studies}

\subsection{Compensation survey}

\begin{lstlisting}
 1  package model
 2  
 3  import (
 4      . "bitbucket.org/dtolpin/infergo/dist"
 5      "bitbucket.org/dtolpin/infergo/mathx"
 6      "math"
 7      "math/rand"
 8  )
 9  
10  type Model struct {Y []bool}
11  
12  type StochasticModel struct {Model}
13  
14  func (m *StochasticModel) Observe(x []float64) float64 {
15      theta := mathx.Sigm(x[0])
16      for i := 0; i != len(m.Y); i++ {
17          coin := rand.Float64()
18          if coin > 0.5 {
19              ll += Flip.Logp(theta, m.Y[i])
20          } else {
21              ll += Flip.Logp(0.5, m.Y[i])
22          }
23      }
24      return ll
25  }
26  
27  type DeterministicModel struct {Model}
28  
29  func (m *DeterministicModel) Observe(x []float64) float64 {
30      theta := mathx.Sigm(x[0])
31      for i := 0; i != len(m.Y); i++ {
32          ll += mathx.LogSumExp(
33              Flip.Logp(theta, m.Y[i])+math.Log(0.5),
34              Flip.Logp(0.5, m.Y[i])+math.Log(0.5))
35      }
36      return ll
37  }
\end{lstlisting}

\clearpage
\subsection{Ball throw}

To keep the deterministic model simple, the player throws the ball with one
of two velocities: either Vw (weak throw) or Vs (strong throw) with equal
probability. The stochastic model can handle any continuous or discrete
random source of velocities.

\begin{lstlisting}
 1  package model
 2  
 3  import (
 4      . "bitbucket.org/dtolpin/infergo/dist"
 5      "bitbucket.org/dtolpin/infergo/mathx"
 6      "math/rand"
 7  )
 8  
 9  const g = 9.80655
10  
11  type Model struct {
12      Vw, Vs float64 // velocities of weak and strong throw
13      L      float64 // distance to the basket
14  }
15  
16  type StochasticModel struct {Model}
17  
18  // The model parameter is Sigm^-1(sin(2*alpha))
19  func (m *StochasticModel) Observe(x []float64) float64 {
20      sin2Alpha := mathx.Sigm(x[0])
21      // Choose a velocity randomly.
22      var v float64
23      if rand.Float64() > 0.5 {
24          v = m.Vs
25      } else {
26          v = m.Vw
27      }
28      d := v * v / g * sin2Alpha
29      return Normal.Logp(m.L, 1, d)
30  }
31  
32  type DeterministicModel struct {Model}
33  
34  // The model parameter is Sigm^-1(sin(2*alpha))
35  func (m *DeterministicModel) Observe(x []float64) float64 {
36      sin2Alpha := mathx.Sigm(x[0])
37      // Compute weighted sum (or integral in continuous case) of
38      // log-probabilities.
39      dw := m.Vw * m.Vw / g * sin2Alpha
40      ds := m.Vs * m.Vs / g * sin2Alpha
41      return 0.5*Normal.Logp(m.L, 1, dw) + 0.5*Normal.Logp(m.L, 1, ds)
42  }
\end{lstlisting}

\clearpage
\subsection{Gaussian mixture model}

\begin{lstlisting}
 1  package model
 2  
 3  import (
 4      . "bitbucket.org/dtolpin/infergo/dist"
 5      "bitbucket.org/dtolpin/infergo/mathx"
 6      "math"
 7      "math/rand"
 8  )
 9  
10  // data are the observations
11  type Model struct {
12      Data  []float64 // samples
13      NComp int       // number of components
14  }
15  
16  type StochasticModel struct {Model}
17  
18  func (m *StochasticModel) Observe(x []float64) float64 {
19      mu := make([]float64, m.NComp)
20      sigma := make([]float64, m.NComp)
21  
22      // Fetch component parameters
23      for j := 0; j != m.NComp; j++ {
24          mu[j] = x[2*j]
25          sigma[j] = math.Exp(x[2*j+1])
26      }
27  
28      // Compute log likelihood under stochastic choices
29      // given the data
30      ll := 0.0
31      for i := 0; i != len(m.Data); i++ {
32          j := rand.Intn(m.NComp)
33          ll += Normal.Logp(mu[j], sigma[j], m.Data[i])
34      }
35  
36      return ll
37  }
38  
39  type DeterministicModel struct {Model}
40  
41  func (m *DeterministicModel) Observe(x []float64) float64 {
42      mu := make([]float64, m.NComp)
43      sigma := make([]float64, m.NComp)
44  
45      // Fetch component parameters
46      for j := 0; j != m.NComp; j++ {
47          mu[j] = x[2*j]
48          sigma[j] = math.Exp(x[2*j+1])
49      }
50  
51      // Compute log likelihood of mixture
52      // given the data
53      ll := 0.0
54      for i := 0; i != len(m.Data); i++ {
55          var l float64
56          for j := 0; j != m.NComp; j++ {
57              lj := Normal.Logp(mu[j], sigma[j], m.Data[i])
58              if j == 0 {
59                  l = lj
60              } else {
61                  l = mathx.LogSumExp(l, lj)
62              }
63          }
64          ll += l
65      }
66  
67      return ll
68  }
\end{lstlisting}
\end{document}